\documentclass[sigconf]{acmart}

\usepackage{amsmath,amsfonts}
\usepackage{algorithmic}
\usepackage{graphicx}
\usepackage{textcomp}
\usepackage{xcolor}

\AtBeginDocument{%
  \providecommand\BibTeX{{%
    \normalfont B\kern-0.5em{\scshape i\kern-0.25em b}\kern-0.8em\TeX}}}

\copyrightyear{2024}
\acmYear{2024}
\setcopyright{acmlicensed}
\acmConference[KDD '24] {Proceedings of the 30th ACM SIGKDD Conference on Knowledge Discovery and Data Mining }{August 25--29, 2024}{Barcelona, Spain.}
\acmBooktitle{Proceedings of the 30th ACM SIGKDD Conference on Knowledge Discovery and Data Mining (KDD '24), August 25--29, 2024, Barcelona, Spain}
\acmISBN{979-8-4007-0490-1/24/08}
\acmDOI{10.1145/3637528.3671534}

\begin{document}
\title{SEFraud: Graph-based Self-Explainable Fraud Detection via Interpretative Mask Learning}

\author{Kaidi Li}
\authornote{Both authors contributed equally to this research and are listed alphabetically.}
\affiliation{
\institution{Huawei Inc}
\city{Shenzhen}
\country{China}}
\email{likaidi@huawei.com}

\author{Tianmeng Yang}
\affiliation{
\institution{Peking University}
\city{Beijing}
\country{China}}
\email{youngtimmy@pku.edu.cn}
\authornotemark[1]

\author{Min Zhou}
\authornote{Corresponding author.}
\affiliation{
\institution{Huawei Inc}
\city{Shenzhen}
\country{China}}
\email{zhoumin27@huawei.com}

\author{Jiahao Meng}
\affiliation{
\institution{Peking University}
\city{Beijing}
\country{China}}
\email{mengjiahao@stu.pku.edu.cn}

\author{Shendi Wang}
\affiliation{
\institution{Huawei Inc}
\city{Shenzhen}
\country{China}}
\email{wangshendi@huawei.com}

\author{Yihui Wu}
\affiliation{
\institution{Huawei Inc}
\city{Shenzhen}
\country{China}}
\email{ngngaifai@huawei.com}

\author{Boshuai Tan}
\affiliation{
\institution{ICBC Limited}
\city{Shanghai}
\country{China}}
\email{tanbs@sdc.icbc.com.cn}

\author{Hu Song}
\affiliation{
\institution{ICBC Limited}
\city{Shanghai}
\country{China}}
\email{songhu@sdc.icbc.com.cn}

\author{Lujia Pan}
\affiliation{
\institution{Huawei Inc}
\city{Shenzhen}
\country{China}}
\email{panlujia@huawei.com}

\author{Fan Yu}
\affiliation{
\institution{Huawei Inc}
\city{Shenzhen}
\country{China}}
\email{fan.yu@huawei.com}

\author{Zhenli Sheng}
\affiliation{
\institution{Huawei Inc}
\city{Shenzhen}
\country{China}}
\email{shengzhenli@huawei.com}

\author{Yunhai Tong}
\affiliation{
\institution{Peking University}
\city{Beijing}
\country{China}}
\email{yhtong@pku.edu.cn}
\authornotemark[2]

\renewcommand{\shortauthors}{Kaidi Li et al.}

\begin{abstract}
  Graph-based fraud detection has widespread application in modern industry scenarios, such as spam review and malicious account detection. While considerable efforts have been devoted to designing adequate fraud detectors, the interpretability of their results has often been overlooked. Previous works have attempted to generate explanations for specific instances using post-hoc explaining methods such as a GNNExplainer. However, post-hoc explanations can not facilitate the model predictions and the computational cost of these methods cannot meet practical requirements, thus limiting their application in real-world scenarios. To address these issues, we propose SEFraud, a novel graph-based self-explainable fraud detection framework that simultaneously tackles fraud detection and result in interpretability. Concretely, SEFraud first leverages customized heterogeneous graph transformer networks with learnable feature masks and edge masks to learn expressive representations from the informative heterogeneously typed transactions. A new triplet loss is further designed to enhance the performance of mask learning. Empirical results on various datasets demonstrate the effectiveness of SEFraud as it shows considerable advantages in both the fraud detection performance and interpretability of prediction results. Specifically, SEFraud achieves the most significant improvement with 8.6\% on AUC and 8.5\% on Recall over the second best on fraud detection, as well as an average of 10x speed-up regarding the inference time. Last but not least, SEFraud has been deployed and offers explainable fraud detection service for the largest bank in China, Industrial and Commercial Bank of China Limited (ICBC). Results collected from the production environment of ICBC show that SEFraud can provide accurate detection results and comprehensive explanations that align with the expert business understanding, confirming its efficiency and applicability in large-scale online services.
\end{abstract}





\maketitle
\section{Introduction}

\begin{figure*}[htbp]
	\centering
        \includegraphics[width=120mm]{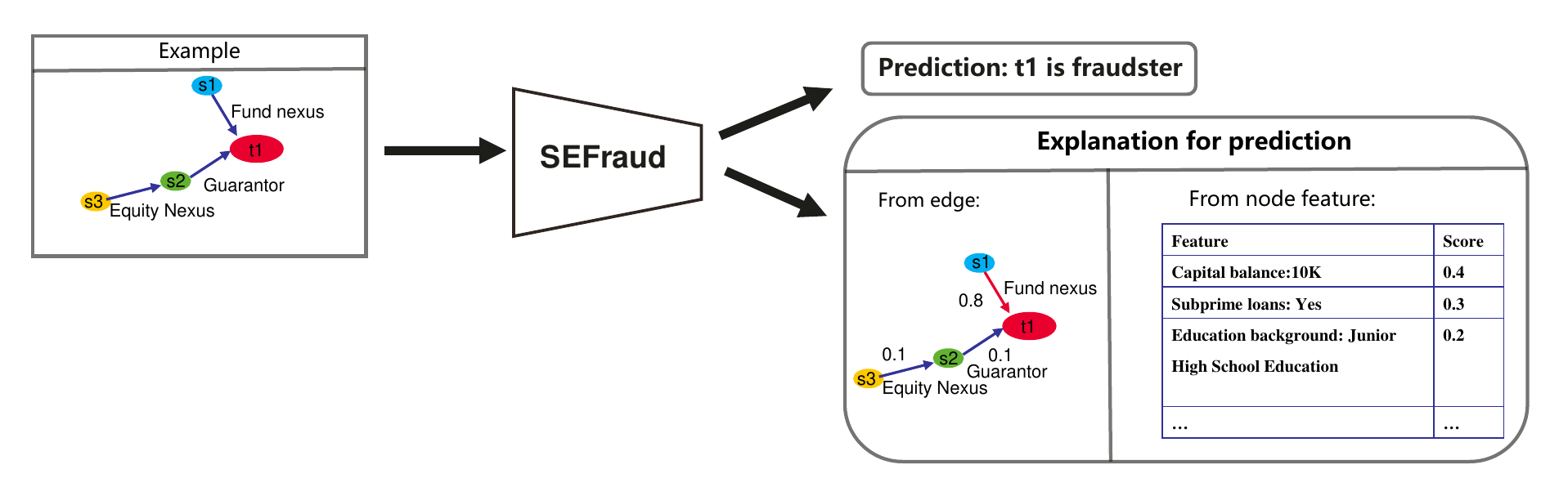}
        \caption{ As a self-explainable model, SEFraud accepts the subgraph which contains both nodes features and edges as the input and provides prediction results and the consistent explanations from edge weight and node feature. For the subgraph of 't1' in the example, the prediction rendered by SEFraud is: 't1' is a fraudster. The explanation for this prediction is twofold: from edge perspective, it is attributed to the fund nexus with user 's1'; from the node feature perspective, the three most important features contributing to the prediction of ‘t1’ as a fraudster are the capital balance, subprime loans, and education background.}
        \label{fig:p_intro}
\end{figure*}

Recent years have witnessed the advances of technology and the prosperity of various internet industry online services~\cite{taher2021commerce,song2023mm,yang2022hrcf}. The connected world brings convenience for society but also fosters potential fraudulent activities~\cite{hilal2022financial,liu2021pick,yang2023mitigating}. Though data mining algorithms and machine learning methods for detecting fraudsters in collections of multidimensional points have been developed over the years, graph machine learning has lately received much attention in the fraud detection community, owing to the success of its applications in various graph-structured data ~\cite{ying2018graph,liu2023mata,yang2021discrete,yang2022graph}. In most fraud detection scenarios, users and fraudsters have rich behavioral interactions when they are engaged in financial activities or post reviews, and such interactions can be represented as graph-like data, which provides practical multifaceted information for identifying fraud behaviors. Aligning with the development of graph machine learning, various graph-based fraud detection methods have been proposed\cite{pourhabibi2020fraud,yang2023mitigating} and applied to various fraud detection scenarios such as malicious account detection~\cite{breuer2020friend,zhong2020financial,liu2019geniepath,liu2018heterogeneous}, anti-money laundry~\cite{li2020flowscope,weber2019anti}, and spam review detection~\cite{dou2020enhancing,zheng2017smoke,deng2022markov}.

Prior research efforts in the realm of graph-based fraud detection have primarily focused on incorporating the development of graph neural networks (GNNs) to identify suspicious fraudsters by capturing relations and connectivity patterns within networks~\cite{liu2019geniepath,dou2020enhancing,liu2021pick}.
For example, GeniePath~\cite{liu2019geniepath} adopts an adaptive path layer for exploring the importance of different-sized neighborhoods and aggregating information from neighbors of different hops. CARE-GNN~\cite{dou2020enhancing} introduces two types of camouflage behaviors of fraudsters and further enhances the GNNs aggregation process with reinforcement learning techniques. Flagging a user or transaction as fraudulent is a complex task. False positives can lead to issues for clients and have a detrimental impact on the platform's credibility. In the realm of fraud detection, there is often an imbalance between positive and negative samples, making it challenging to identify the latter. To remedy the class imbalance problem and improve the identification ability for fraudsters, a Pick and Choose Graph Neural Network (PC-GNN )~\cite{liu2021pick} is proposed. The label-balanced sampler is designed to pick nodes and edges for the sub-graph training process, with a neighborhood sampler to pick neighbors. 

However, these approaches are trained end-to-end in a black-box manner, thus lacking transparency and interpretability for the detection results and limiting their practical applications where high confidence and explanations are required, such as financial fraud detection.
Moreover, examining and interpreting identified negative samples requires substantial human resources. The methods mentioned above face practical challenges due to the limited transparency and interpretability of the predicted outcomes. xFraud~\cite{rao2020xfraud} partly addressed this problem and combined a detector with an explainer, such as GNNExplainer~\cite{ying2019gnnexplainer}, to provide explanations for detection results.
Nevertheless, GNNExplainer needs to be retrained for every single instance; thus, it is time consuming and needs to improve efficiency when explaining large-scale graph nodes. While PGExplainer~\cite{luo2020parameterized} improves the efficiency and accuracy compared to the GNNExplainer,  it primarily focuses on the edge perspective, thereby overlooking the explanations for node features. These node features are crucial in the prediction process of fraud detection models. In addition, both GNNExplainer~\cite{ying2019gnnexplainer} and PGExplainer~\cite{luo2020parameterized} can only provide post-hoc explanations, which can not facilitate the model predictions.

To facilitate detector models with recognizing important node features and topology edges for higher fraud detection accuracy, as well as providing explanations more efficiently, we propose a new graph-based \textbf{S}elf-\textbf{E}xplainable \textbf{Fraud} detection method termed as SEFraud, as depicted in Figure \ref{fig:p_intro}. Concretely, we leverage a customized heterogeneous graph transformer with a feature attention net and edge attention net, and train the model in a unified manner. Besides, we hope the learned masks can maximize the identification of essential edges and node features; if the weights are reversed, the results will also be drastically interfered. To this end, a supervised contrastive triplet loss is designed to enhance the mask learning. Empirical studies on various datasets demonstrate that SEFraud significantly improves both the fraud detection performance and interpretability of prediction results. Ablation and case studies are also provided to evaluate the proposed methods thoroughly.

In summary, the advantages of SEFraud are three folds:
\begin{itemize}
    \item \textbf{Higher fraud detection performance.} 
    Through the learnable masking module, SEFraud re-weights the initial node features and imposes constraints on the influence of edges during the forward process, which enables SEFraud to more effectively capture essential node features and fundamental relations, thereby enhancing the performance of the fraud detection model.
    
    \item \textbf{Milliseconds response efficiency.} 
    Rather than retraining the explanation network for each new instance, the learned feature mask and edge mask are self-explanatory, which allows SEFraud to generate precise and comprehensive explanations within milliseconds. 
    
    \item \textbf{Excellent application effect.} SEFraud has been deployed and offers explainable detection service for the largest bank in China, Industrial and Commercial Bank of China Limited (ICBC). Results on the real production environment with real graphs from ICBC prove that SEFraud can provide both accurate detection results and explanations that align with their business understanding, showing its efficiency and applicability in practice.
    
\end{itemize}
\section{Preliminaries and Related Work}
 Graph-based fraud detection can be applied to a wide range of scenarios, such as financial transactions~\cite{ma2018graphrad}, human behaviours~\cite{branting2016graph}, and social media interactions~\cite{wang2019fdgars}, to uncover patterns and anomalies that indicate fraudulent activity. In this section, we introduce the basic graph-based fraud detection task definition and review the recent developments on GNNs and their heterogeneous variants. Then, we briefly introduce works about GNNs and interpretability and their applications in the fraud detection scenario.

\subsection{Problem Formulation}
Given a graph $\mathcal{G=(V,E,X,Y)}$ with node set $\mathcal{V}$ and edge set $\mathcal{E}$; $x_i \in \mathcal{X}$ represents a $d$-dimension feature vector of node $v_i$ and $x_i \in R^d$; $\mathcal{Y}$ is the set of labels for each node in $\mathcal{V}$. Generally, graph-based fraud detection can be viewed as a binary node classification problem on $\mathcal{G}$ to classify the nodes into benign ($y_v=0$) or fraudulent ($y_v=1$) groups. 

It is noted that if there are one type of node and various types of relations in  $\mathcal{G}$, a multi-relation graph can be defined as $\mathcal{E} = \{E_r\}|_r^R$ with $R$ different types of relations. Considering a heterogeneous scenario that may contain different types of nodes and different types of relations simultaneously, we further define a heterogeneous graph to be associated with a node type mapping function $\mathcal{\tau: V \rightarrow A }$ and an edge type mapping function $\mathcal{\phi: E \rightarrow R}$. $\mathcal{A}$ and $\mathcal{R}$ denote the sets of predefined node types and relation types, and $\mathcal{|A| + |R|} > 2$.

\subsection{GNNs for Fraud Detection}
GNNs are widely used for fraud detection by analyzing the connections and relationships between entities in a network~\cite{9565320}. Most existing GNNs~\cite{kipf2016semi,velivckovic2017graph,hamilton2017inductive} adopt message-passing framework~\cite{gilmer2017neural}, which applies local aggregation to learn node representations. 
At each propagation step $t$, the hidden representation of node $v$ is derived by:
\begin{align}
    h_v^t = f(h_v^{t-1}, {\rm aggr} (h_u^{t-1} | u \in N(v))),
\end{align}
where $h_v^0 = x_v$, $\rm aggr(\cdot)$ denotes a differentiable, permutation invariant function to aggregate neighbor information, $f(\cdot)$ is a transformation function between two propagation steps, $N(v)$ is the connected neighbors of $v$.

Many previous works conduct semi-supervised node classification tasks on single-relation graphs~\cite{liu2019geniepath} or define multi-relation graph ~\cite{dou2020enhancing,wang2019semi,liu2021pick,liu2020alleviating} to tackle fraud detection tasks. The common insight is to design different massage passing mechanisms for aggregating neighborhood information. For example, GeniePath~\cite{liu2019geniepath} learns convolutional layers and neighbor weights using LSTM~\cite{hochreiter1997long} and the attention
mechanism~\cite{velivckovic2017graph}. SemiGNN~\cite{wang2019semi} applies a GNN-based hierarchical attention mechanism to detect fraudsters on Alipay. GraphConsis~\cite{liu2020alleviating} and CARE-GNN ~\cite{dou2020enhancing} filter dissimilar neighbors before aggregation to discover camouflage
fraudsters. PC-GNN~\cite{liu2021pick} identifies and solves the label imbalance issue by node resampling. Heterogeneous GNNs~\cite{hu2020heterogeneous,wang2019heterogeneous,fu2020magnn,yang2023simple} are capable of learning embeddings for heterogeneous graphs but still remain underexplored for fraud detection scenarios.

\subsection{Explainability of GNNs} 

Despite the notable achievements of graph-based methods, their lack of transparency hinders easy comprehension of the predictions. Nevertheless, enhancing explanations and providing the model credibility which offers valuable guidance to business professionals in terms of prevention and control holds significant importance, especially for applications such as fraud detection.
Recently, several models have been proposed to explain the predictions of GNNs, such as XGNN~\cite{yuan2020xgnn}, GNNExpaliner~\cite{ying2019gnnexplainer}, and PGExplainer~\cite{luo2020parameterized}. GNNExplainer~\cite{ying2019gnnexplainer} learns soft masks for edges and node features
to explain the predictions via mask optimization. PGExplainer~\cite{luo2020parameterized} learns approximated discrete masks for edges to explain the predictions via the reparameterization trick. XGNN~\cite{yuan2020xgnn} trains a graph generator to provide model-level explanations and is only used for graph classification models.
Few works pay efforts to provide explanations for graph-based fraud detection. xFraud~\cite{rao2020xfraud} combines the fraud detector with an explainer using GNNExplainer and provides explanations to facilitate further processes. However, these methods can only provide post-hoc explanations. Besides, GNNExpaliner needs to re-train the explanation network for each new instance and thus is criticized for its low efficiency. While PGExplainer~\cite{luo2020parameterized} enhances both the accuracy and efficiency of explanations compared to GNNExplainer~\cite{ying2019gnnexplainer}, it still has limitations. Specifically, PGExplainer’s explanations are solely from the perspective of edge aspects, leading to a certain degree of incompleteness. This becomes particularly problematic in scenarios that necessitate a comprehensive explanation encompassing both node features and relations simultaneously. By incorporating a self-explanatory mask learning module into the GNN model, we aim to improve representation learning of the GNN detector to enhance the model's prediction ability, as well as generate high-quality and comprehensive explanations with milliseconds response efficiency.
\section{The Proposed Method}
\begin{figure*}[htbp]
	\centering
        \includegraphics[width=\linewidth]{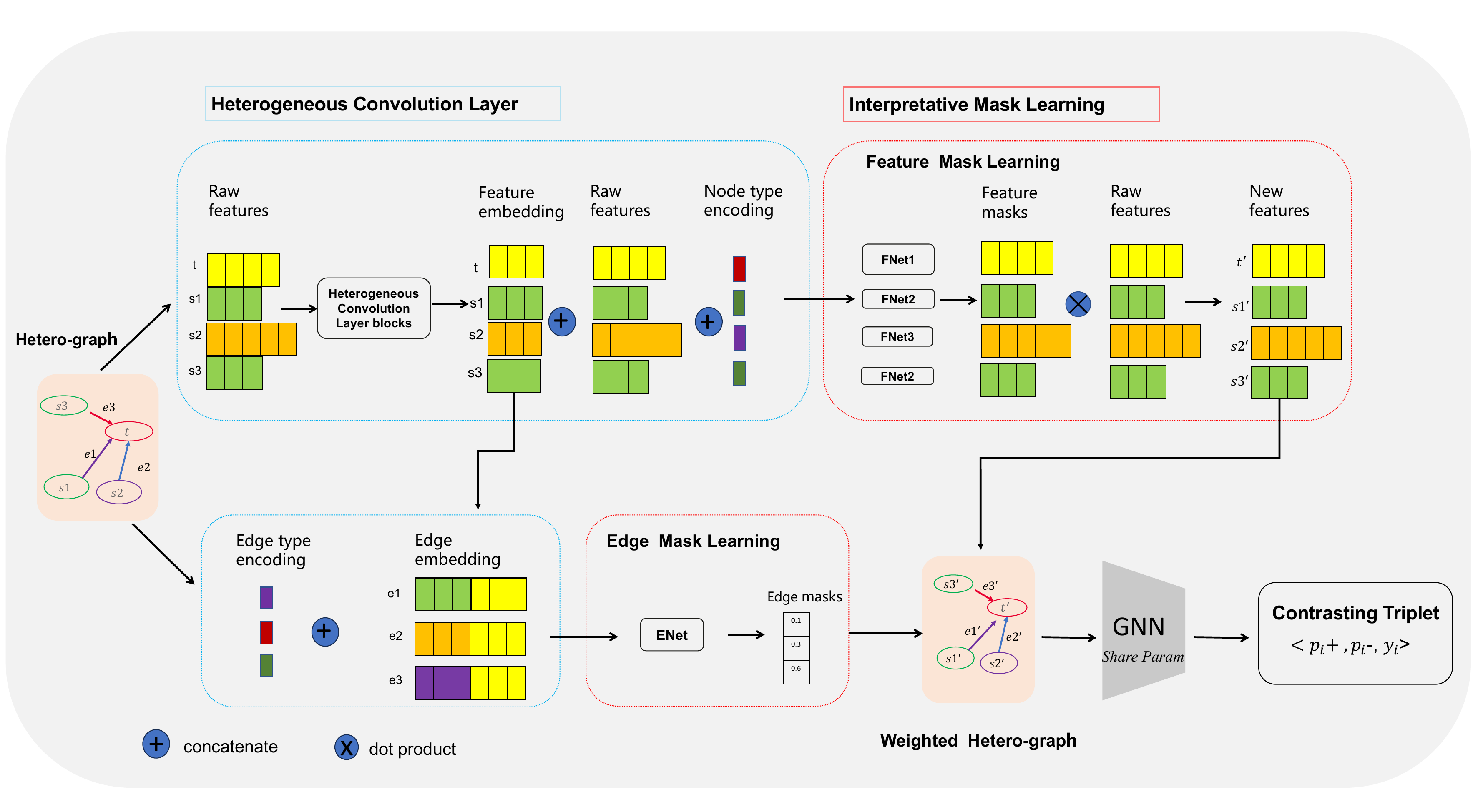}
        \caption{The architecture of SEFraud. A heterogeneous convolution layer is utilized to aggregate the hetero-graph information and generate the feature embedding for each node. These feature embeddings, raw features, and node type encodings for each node are then concatenated to form the input for FNet. An edge embedding consists of the node embeddings at its two ends, and concatenates with the edge type encodings to form the the input for ENet. The learned feature masks and edge masks are further leveraged to reconstruct a weighted hetero-graph, which serves as the input for the GNN/Detection model. A contrastive triplet loss is then constructed based on the output of the model for the training process.}
        \label{fig:arch}
\end{figure*}

In this section, we present the proposed Self-Explainable Fraud framework. As sketched in Figure \ref{fig:arch},  SEFraud first utilizes customized heterogeneous graph transformer networks to extract meaningful representations from diverse transaction types, which are then incorporated by the learnable feature masks and edge masks, generating the reweighted graph. A novel triplet loss is specifically designed to enhance the mask learning. We detail the information aggregation and interpretative mask learning in Section \ref{hgl}  and \ref{iml}, respectively. Then we explain how to aggregate information from the reweighted graph and construct the contrasting triplet in Section \ref{ctl}.

\subsection{Heterogeneous Convolution Layer}
\label{hgl}
Considering various node types and edge types in real-world industry scenarios, we utilize a Heterogeneous Graph Transformer (HGT)~\cite{hu2020heterogeneous} to construct our convolution layer, and aggregate heterogeneous neighborhood information from source nodes to get a contextualized representation for target node. The process can be decomposed into three components: Heterogeneous Mutual Attention, Heterogeneous Message Passing, and Target-Specific Aggregation. Denoting the output of the $l-th$ layer is $H^l$, we can stack multiple layers to learn the node representations as follows:
\begin{align}
    H^{l+1} = \mathop{Aggregate}\limits_{\forall s \in N(t), \forall e \in E(s,t)}(\mathop{Attention}(s,t) \cdot \mathop{Message}(s)),
    \label{Eq:aam}
\end{align}
where $N(t)$ is the neighborhood nodes of target node $t$, $E(s,t)$ is the edges connecting source node $s$ and target node $t$.

\subsubsection{Heterogeneous Mutual Attention}
Similar to the vanilla Transformer~\cite{vaswani2017attention}, HGT maps target node $t$ into a Query vector, source node $s$ into a Key vector, and calculate their dot product as attention. Considering $h$ attention heads, the calculation is be formulated as:
\begin{gather}
    \mathop{Attention}(s,e,t) = \mathop{Softmax}\limits_{s \in N(t)}(\mathop{\Vert}\limits_{i \in [1,h]} ATT\text{-}Head^i(s,e,t)), \\
    ATT\text{-}Head^i(s,e,t) = K^i(s)W_{\phi(e)}^{ATT}Q^i(t), \\
    K^i(s) = K\text{-}Linear_{\tau(s)}^i(H^{l-1}[s]), \\
    Q^i(t) = Q\text{-}Linear_{\tau(t)}^i(H^{l-1}[t]),
    \label{Eq:attention}
\end{gather}
where $Linear$ means linear projection and $\tau(\cdot)$ represent the node types.

\subsubsection{Heterogeneous Message Passing}
To incorporate the meta relations of different types, the multi-head $Message$ is further calculated by applying different edge type projections as:
\begin{gather}
    \mathop{Message}(s,e,t) = \mathop{\Vert}\limits_{i \in [1,h]} MSG\text{-}Head^i(s,e,t), \\
    MSG\text{-}Head^i(s,e,t) = M\text{-}Linear_{\tau(s)}^i(H^{l-1}[s]) W_{\phi(e)}^{MSG}. 
    \label{Eq:msg}
\end{gather}

\subsubsection{Target-Specific Aggregation}
In this step, the attention vector and corresponding message from source node $s$ are combined to get the neighborhood information as:
\begin{gather}
    \hat{H}^{l}[t] = \sum\limits_{s \in N(t)} \mathop{Attention}(s,e,t) \cdot \mathop{Message}(s,e,t). 
    \label{Eq:msg}
\end{gather}

Finally, target-specific linear projection are applied to $\hat{H}^{l}[t]$  and get the updated vector $H^{l}[t]$ with a residual connection:
\begin{gather}
    H^{l}[t] = T\text{-}Linear_{\phi(t)}(\sigma(\hat{H}^{l}[t])) + H^{l-1}[t],
    \label{Eq:msg}
\end{gather}
where $\sigma$ is the activation function.

\subsection{Interpretative Mask Learning}
\label{iml}
Interpretability in graph-based fraud detection tasks includes two aspects. Firstly, influential edges offer topology explanations for correlations between nodes. Secondly, certain critical features provide the properties of nodes to be flagged as fraudsters. By stacking $L$ layers, we can obtain the node representations of the entire graph, denoted as $H(L)$.
Unlike the attention weights in HGT that vary across different layers, we aim to learn a unified, consistent feature mask and edge mask among all aggregating layers. 

In order to achieve this, we integrate a feature attention network (FNet) and an edge attention network (ENet) following the heterogeneous convolution layers. Given the heterogeneity of node types and edge types, we also implement type encoding for each node type and edge type.
Specifically, for the feature mask, we concatenate the initial feature $H^0$ , the aggregated feature $H^L$, and the node type encoding of each node. We then utilize a $FNet$ for different node types to learn the node mask. For the edge mask, we concatenate the source node and target node features along with their edge type encodings. We then employ an $ENet$ to learn the edge mask. 
Formally, the learned feature mask $SE\text{-}Mask_{n}$ and edge mask $SE\text{-}Mask_{e}$ are acquired by:
\begin{gather}
    SE\text{-}Mask_{n} = \mathop{FNet}\limits_{\forall v \in N}(H^0 \Vert H^L \Vert NTE(\tau(v)), \\
    SE\text{-}Mask_{e} = \mathop{ENet}\limits_{\forall e \in (s,e,t))}(H^L[s] \Vert H^L[t] \Vert ETE(\phi(e)),
    \label{Eq:msg}
\end{gather}
where $\mathop{FNet(\cdot)}$ and  $\mathop{ENet(\cdot)}$ are Multi-Layer Perceptrons, $NTE(\cdot)$ and $ETE(\cdot)$ are node type encoding function and edge type encoding function, respectively.

\subsection{Contrastive Triplet Loss}
\label{ctl}
With the learned feature mask and edge mask, the initial node features are re-weighted, and the influence of edges is constrained during the forward process of the model. Intuitively, the model's prediction results with learned masks should align closely with the ground truth. Conversely, if we assign negative weights to the masks, the model's prediction results should diverge significantly. Denoting the positive prediction results of node $i$ as $p_i+$, the negative results as $p_i-$, and the ground truth as $y_i$, we have designed a contrastive triplet loss function, denoted as $\mathcal{L}_{tr}$, that operates on each triplet pair $<y_i,p_i+,p_i->$. This function penalizes the distance between the positive pair and the negative pair with a margin $\alpha$:
\begin{gather}
    \mathcal{L}_{tr} = \sum\limits_{i \in V_l}max(0,dis(y_i,p_i+)-dis(y_i,p_i-)+\alpha),
    \label{Eq:trloss}
\end{gather}
where the cross-entropy are utilized as the $dis(\cdot)$ metric and $V_l$ is the training set.

\subsection{SEFraud for Detection and Explanation}
Finally, we construct the SEFraud framework for fraud detection tasks, incorporating the aforementioned designs. Besides,  the framework provides explanations through the learned feature mask and edge mask. 
The accuracy of fraudster classification is guaranteed with a cross-entropy loss, denoted as $\mathcal{L}_{ce}$:
\begin{gather}
    \mathcal{L}_{ce} = \sum\limits_{i \in V_l}y_i\log{p_i+},
    \label{Eq:celoss}
\end{gather}

 To enhance both the effect of classification and mask learning, we train the model with a combination of the cross-entropy classification loss $\mathcal{L}_{ce}$ and the contrasting triplet loss $\mathcal{L}_{tr}$. The combined loss function is defined as follows:
\begin{gather}
    \mathcal{L} = (1-\lambda) \cdot \mathcal{L}_{ce} + \lambda \cdot \mathcal{L}_{tr},
    \label{Eq:total loss}
\end{gather}
where the $\lambda$ is a hyperparameter to balance fraudster classification and interpretive mask learning. 
\section{Experiments}
For a comprehensive evaluation of SEFraud, we conduct experiments for both fraud detection and interpretation tasks, and compare it with various baselines. An ablation study is also performed to show the effectiveness of our designs.
Case studies, time efficiency comparison, and deployment effect are further provided to demonstrate the advantages of our method in industry application scenarios.

\subsection{Datasets}
The datasets include three fraud detection datasets and four widely used explanation datasets, with the statistics summarized in Table~\ref{tab:datasets} and Table~\ref{tab:datasets2}.
\subsubsection{Fraud detection datasets} 
\begin{itemize}
    \item \textbf{Yelp} and \textbf{Amazon}\cite{dou2020enhancing}. The Yelp dataset includes hotel and restaurant reviews filtered (spam) and recommended (legitimate) by Yelp. The nodes in the graph are reviews with three relations: the reviews posted by the same user, the reviews under the same product with the same star rating, and the reviews under the same product posted in the same month. The Amazon dataset includes product reviews under the Musical Instruments category. The nodes in the graph of the Amazon dataset are users with 100-dimension features and also contain three relations: users reviewing at least one same product, users having at least one same star rating within one week, and users with top-5\% mutual review TF-IDF similarities. The users with more than 80\% helpful votes are benign entities and users with less than 20\ helpful votes are fraudulent entities. Handcrafted features are extracted from prior works~\cite{rayana2015collective,zhang2020gcn} as the raw node features for the datasets, respectively. 
    
    \item \textbf{ICBC}. The financial fraud detection dataset, termed ICBC, is generated by TabSim~\footnote{TabSim is open-source and can be found at the following link: https://gitee.com/mindspore/xai/wikis/TabSim}, a tool based on the statistical characteristics of debts and customers provided by ICBC (Industrial and Commercial Bank of China), one of the largest banks in China. The nodes of ICBC data represent customers and their debts, each containing 24 features, including profession, education background, and various asset metrics. The edges depict relations between nodes, encompassing five distinct types: equity association, card holder, against pledge, staff, and others.
\end{itemize}

\begin{table}[t]
    \caption{Statistics of the fraud detection datasets.}
    \label{tab:datasets}
    \centering
    \small
    \setlength{\tabcolsep}{4.5mm}{
    \begin{tabular}{lccc}
    \toprule
    \textbf{Fraud Detection} & Yelp & Amazon & ICBC\\
    \midrule
    \#Nodes & 45954 & 11944 & 100000 \\
    \#Features & 32 & 25 & 24 \\
    \#Edges & 3846979 & 4398392 & 102472 \\
    \#Edge Types & 3 & 3 & 5 \\
    \#Classes & 2 & 2 & 2 \\
    \bottomrule
    \end{tabular}
    }
\end{table}

\begin{table}[t]
    \caption{Statistics of the interpretation datasets.}
    \label{tab:datasets2}
    \centering
    \small
    \setlength{\tabcolsep}{1.5mm}{
     \begin{tabular}{lcccc}
    \toprule
    \textbf{Interpretation} & {BA-2motifs}  & {BA-Shapes} & {Tree-Cycles} & {Tree-Grids}\\
    \midrule
    \#Graphs & 1000 & 1 & 1 & 1\\
    \#Nodes & 25000 & 700 & 871 & 1231\\
    \#Edges & 51392 & 4110 & 1950 & 3410\\
    \#Labels & 2 & 4 & 2 & 2\\
    \bottomrule
    \end{tabular}
    }
\end{table}

\subsubsection{Interpretation datasets} 
\begin{itemize}
    \item \textbf{BA-Shapes}, \textbf{Tree-Cycles} and \textbf{Tree-Grids}. These datasets are synthetic graphs for node-level explanation. BA-Shapes is a single graph consisting of a base Barabasi-Albert (BA) graph with 300 nodes and 80 “house”-structured motifs. These motifs are attached to randomly selected nodes from the BA graph. After that, random edges are added to perturb the graph. Nodes in the base graph are labeled with 0; the ones locating at the top/middle/bottom of the “house” are labeled with 1,2,3, respectively. For the Tree-Cycles dataset, an 8-level balanced binary tree is adopted as the base graph. An 80 six-node cycle motifs are attached to randomly selected nodes from the base graph. Tree-Grid is constructed similarly to Tree-Cycles, except that 3-by-3 grid motifs are used to replace the cycle motifs.
    \item \textbf{BA-2motifs}. For graph level explanation, BA-2motifs adopts the BA graphs as base graphs. Half of the graphs are attached with “house” motifs, and the rest are attached with five-node cycle motifs. Graphs are assigned to one of 2 classes according to the type of attached motifs.
\end{itemize}

\subsection{Baselines}
To verify the ability of SEFraud in graph-based fraud detection tasks, we compare it with various GNN baselines under the semi-supervised learning setting, including:
\begin{itemize}
   \item  Four representative general GNN models for node classification: GCN~\cite{kipf2016semi}, GAT~\cite{velivckovic2017graph}, GraphSAGE~\cite{hamilton2017inductive} and RGCN~\cite{schlichtkrull2018modeling}.
    \item Two strong graph-based fraud detection methods without explainability: GeniePath~\cite{liu2019geniepath} and CARE-GNN~\cite{dou2020enhancing}.
    \item An explainable fraud prediction system xFraud~\cite{rao2020xfraud} that consists of a heterogeneous GNN detector and a post-hoc explainer.
\end{itemize}

Among those baselines, GCN, GAT, GraphSAGE, and GeniePath are built on homogeneous graphs. Thus, all relations are merged together. RGCN and CARE-GNN run on multi-relation graphs, while xFraud and SEFraud can handle information from different relations and node types based on the heterogeneous graph convolution.

\subsection{Experimental Settings}
To make a fair comparison, we closely follow the experimental procedure with~\cite{dou2020enhancing} on fraud detection and~\cite{ying2019gnnexplainer,luo2020parameterized} on explanation. We use the same feature vectors, labels, and train/val/test splits for all baselines. For SEFraud, the layers of heterogenous convolution $L$ is 2, number of heads $h$ is 4, margin of contrastive triplet loss $\alpha$ is 0.1, trade off parameter of combined loss function $\lambda$ is 0.3. Other settings are tuned from: hidden size $\in \{8,16,32,64\}$, learning rate $\in \{5e\text{-}3,1e\text{-}2\}$, and L2 regularization weight $\in \{1e\text{-}3,5e\text{-}4\}$. We implement our methods using Pytorch Geometric~\cite{Fey/Lenssen/2019pyg} with the Adam optimizer~\cite{kingma2014adam}.

\begin{table*}[t]
    \caption{Fraud detection performance comparasion. The best result on per benchmark is highlighted.}
    \label{tab:experiment}
    \renewcommand\arraystretch{1}
    \centering
    \small
    \setlength{\tabcolsep}{6mm}{
    \begin{tabular}{l|cc|cc|cc}
    \toprule
    \textbf{Dataset} & \multicolumn{2}{c|}{Yelp} & \multicolumn{2}{c|}{Amazon} & \multicolumn{2}{c}{ICBC}\\
    \midrule
    Metrics & AUC & Recall & AUC & Recall & AUC & Recall\\
    \midrule
    GCN & 52.47 & 50.81 & 74.34 & 67.45 & 50.72 & 50.53\\
    GAT & 56.24 & 54.52 & 75.16 & 65.51 & 87.70 & 79.22\\
    GraphSAGE & 54.00 & 52.86 & 75.27 & 70.16 & 93.93 & 93.02\\
    RGCN & 53.38 & 50.43 & 74.68 & 67.68 & 92.42 & 91.29\\
    GeniePath & 55.91 & 50.94 & 72.65 & 65.41 & 99.27 & 94.48\\
    CARE-GNN & 75.70 & 71.92 & 89.73 & 88.48 & 98.96 & 97.18 \\
    XFraud & 79.88 & 72.46 & 92.49 & 88.26 & 95.68 & 95.42 \\
    \midrule
    SEFraud(ours) & \textbf{86.77} & \textbf{78.64} & \textbf{93.23} & \textbf{88.67}  & \textbf{99.69} & \textbf{99.38}  \\
    \midrule
    SEFraud(\textit{w/o triplet loss}) & 83.51 & 74.23 & 92.47 & 86.23 & 96.68 & 96.37 \\
    SEFraud(\textit{w/o node mask}) & 86.21 & 78.29 & 93.16 & 87.16 & 96.97 & 96.36 \\
    SEFraud(\textit{w/o edge mask}) & 84.06 & 75.91 & 92.64 & 85.47 & 99.37 & 98.17\\
    \bottomrule
    
    \end{tabular}
    }
\end{table*}

\subsection{Fraud Detection Performance}
In most fraud detection scenarios, the nodes are predominantly benign, with the proportion of fraudulent entities being significantly low. From the application perspective, the task of fraud detection is supposed to pay more attention to recognizing potential fraudsters. Thus, we report the ROC-AUC (AUC) and Recall metrics following~\cite{dou2020enhancing} to evaluate the overall performance of all methods. AUC is computed based on the relative ranking of prediction probabilities of all instances, which could eliminate the influence of imbalanced classes. Recall measures how the model correctly identifies positive instances (true positives) from all actual positive samples, thus representing the ability to flag potential fraudsters.

Table~\ref{tab:experiment} summarizes the prediction results of fraud detection among three datasets. Most GNNs built on homogeneous graphs such as GCN and GeniePath, or simply aggregate information from different relations such as RGCN, show inferior performance compared with methods that can handle complex multi-relation graphs (CARE-GNN) or apply heterogeneous graph convolution (xFraud and SEFraud) on Yelp and Amazon. In particular, xFraud achieves remarkable improvements on these two datasets. It is worth mentioning that GeniePath outperforms other baselines on the ICBC dataset, which may be because ICBC is more sparse, and it is critical to explore essential neighborhoods. In a sparse and noisy scenario, GeniePath and CARE-GNN, which are capable of adaptively filtering neighborhood information, can even overpass xFraud. 

On the one hand, SEFraud tackles heterogeneity compared with general GNNs. On the other hand, compared with xFraud, SEFraud can enhance the heterogeneous convolution process by learning feature mask and edge mask, thus consistently outperforming all baselines on both AUC and Recall metrics. Specifically, SEFraud achieves an 8.6\% improvement on AUC and an 8.5\% improvement on Recall over xFraud on the Yelp dataset.

We further conducted ablation studies to evaluate the effectiveness of the each component in the proposed SEFraud. Specifically, we denoted models without a specific component using \textit{w/o} and re-evaluate the performance. The results presented in the bottom part of Table~\ref{tab:experiment} demonstrate that removing any of the components leads to performance degradation, confirming the efficacy of each module in the proposal. Notably, for denser graphs like Yelp and Amazon, the edge masks has a more significant impact than the feature masks as considerable degradation on the performance is observed if the corresponding module is dropped. Conversely, in the case of ICBC where edges are sparse, the node mask plays a critical role. 
Additionally, the triplet loss is essential for learning high-quality masks and exhibits a noticeable decrease on performance when removed.

\subsection{Interpretability}
Given that the FNet and ENet are trained alongside the detection model, a higher level of model prediction accuracy demonstrates the validity of the learned masks applied to the original graphs. The learned masks reflect the significance of the node features and edges. Therefore, they are interpretative and can directly be utilized to elucidate the model predictions.

\subsubsection{Quantitative evaluation} To quantitatively verify the interpretability of the learned masks, we follow the setting in GNNExplainer~\cite{ying2019gnnexplainer} and PGExplainer~\cite{luo2020parameterized} to construct experiments on the synthetic datasets on node classification and graph classification tasks. Concretely, we incorporate our proposed mask learning paradigm (termed as SE-Mask) into a GCN backbone used in GNNExplainer and PGExplainer, and leverage the learned edge masks as interpretation when GCN training is finished.

\begin{table}[tbp]
    \caption{Interpretation Performance (AUC).}
    \label{tab:explain}
    \renewcommand\arraystretch{1}
    \centering
    \small
    \scalebox{1.0}{
    \begin{tabular}{l|c|c|c|c}
    \toprule
    \textbf{} & {BA-2motifs} & {BA-Shapes} & {Tree-Cycles} & {Tree-Grids}\\
    \midrule
    GRAD & 71.7 & 88.2 & 90.5 & 61.2\\
    ATT & 67.4 & 81.5 & 82.4 & 66.7\\
    Gradient & 77.3 & - & - & -\\
    GNNExplainer & 74.2 & 92.5 & 94.8 & 87.5\\
    PGExplainer & 92.6{\tiny$\pm$2.1} & 96.3{\tiny$\pm$1.1} & \textbf{98.7{\tiny$\pm$0.7}} & 90.7{\tiny$\pm$1.4}\\
    \midrule
    SE-Mask & \textbf{97.8{\tiny$\pm$1.6}} & \textbf{99.8\tiny{$\pm$0.1}} & 96.5{\tiny$\pm$1.3} & \textbf{95.8{\tiny$\pm$2.7}}\\
    \bottomrule
    \end{tabular}
    } 
\end{table}

The explanation problem is formalized as a binary classification task, where edges in the ground-truth explanation are treated as labels, and importance weights given by the explanation method are viewed as prediction scores. A suitable explanation method assigns higher weights to edges in the ground truth motifs than outside ones. Thus, AUC is adopted as the metric for quantitative evaluation.
Results in Table~\ref{tab:explain} show that our proposed SE-Mask achieves remarkable interpretation performances compared with baseline explanation methods in three datasets. For Tree-Cycles, SE-Mask performs inferior to PGExplainer but still overpasses GNNExpalainer. The average lifts on the AUC of SE-Mask over GNNExpalainer and PGExplainer are 12.8\% and 3.2\%, demonstrating its superior interpretability. 

\begin{figure}[t]
\centering
\includegraphics[width=0.48\textwidth]{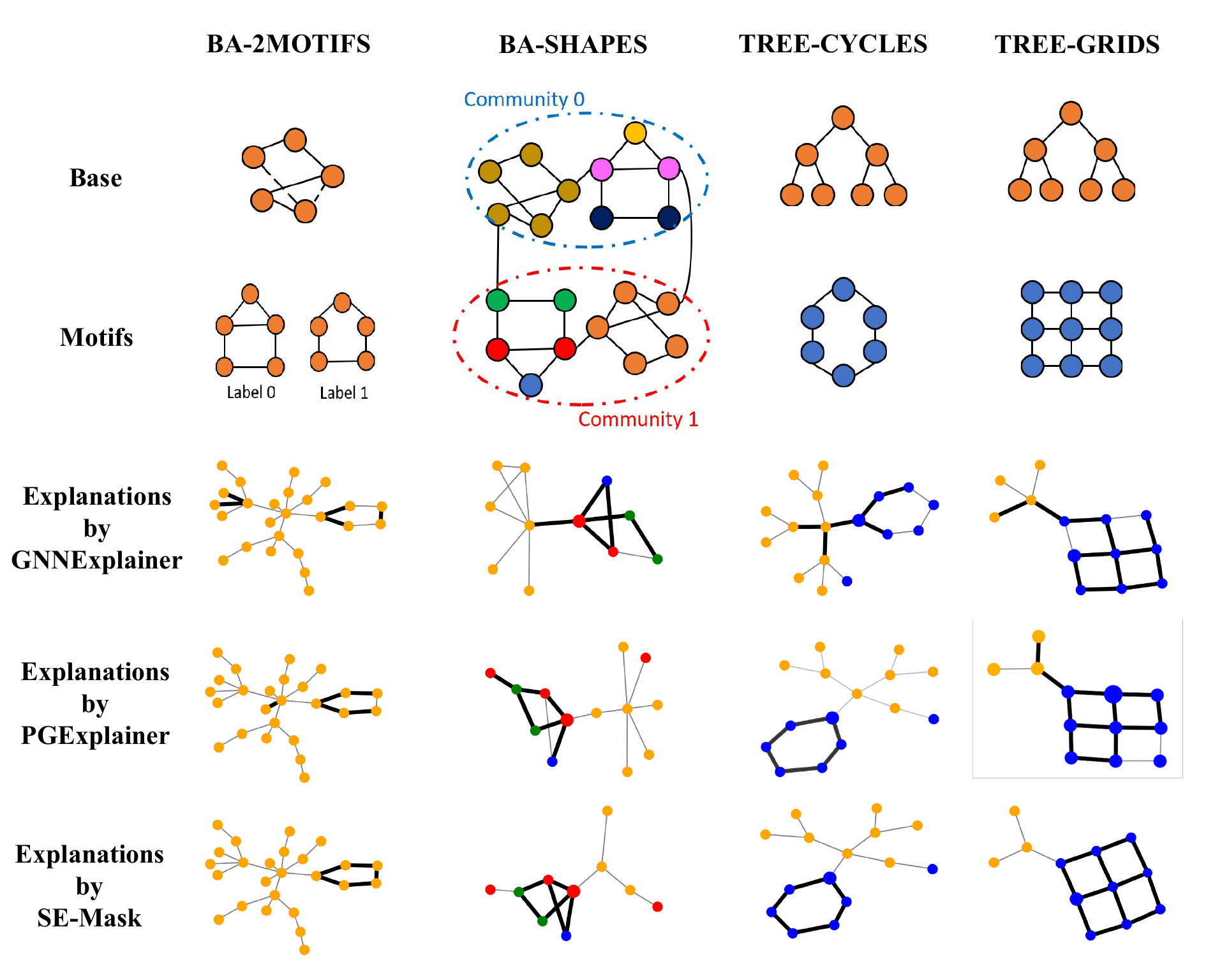}
\caption{Qualitative explanation examples comparison. Node labels are represented by their colors. Explanations of instance in each dataset are highlighted by bold black edges ranked by their importance weights.}
\label{fig:qualitative}
\end{figure}

\subsubsection{Qualitative evaluation} 

We also choose an instance for each dataset and visualize its explanations given by SE-Mask and baselines in Figure \ref{fig:qualitative}. We first compute the edge weights for each instance and highlight the top-K edges where K is the number of edges in each explanation motif. For both graph classification and node classification tasks, SE-Mask can accurately identify edges inside the explanation motifs (i.e., the "house," "cycle," and "3-by-3 grid"), while GNNExplainer and PGExplainer have confused some essential edges. These visualizations demonstrate that SE-Mask possesses a more robust ability for explanation.

\begin{figure}[tbp]
\centering
\includegraphics[width=0.48\textwidth]{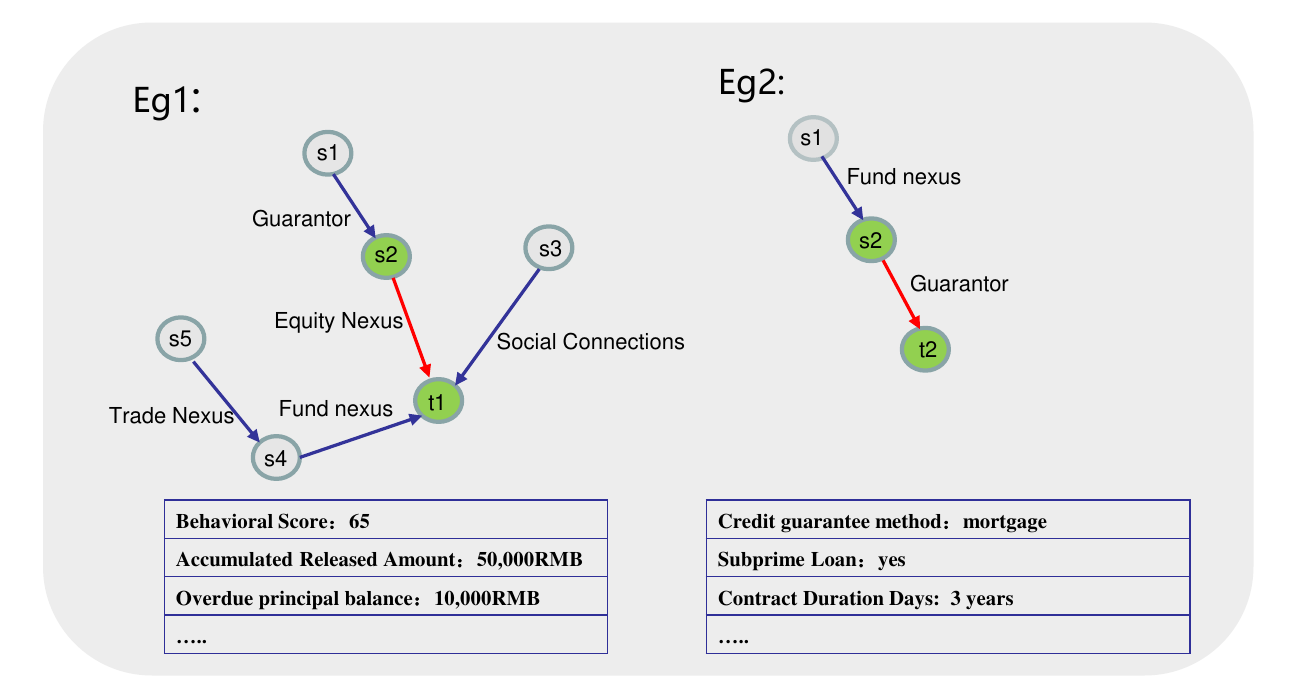}
\caption{Explanation examples from ICBC dataset.}
\label{fig:case_study}
\end{figure}

To further validate the credibility of our explanations on financial fraud detection, we randomly selected 100 nodes that were predicted to be fraudsters, which were further manually checked and confirmed by business experts of ICBC. The experts corroborated that the explanations provided by our system, SEFraud, are not only logical but also congruent with their extensive business acumen. This validation underscores the credibility of our system in the realm of financial fraud detection. Two detailed examples are sketched in Figure \ref{fig:case_study}.

For node prediction `t1',  from the perspective of edge analysis, the most pivotal edge contributing to its classification as a fraudulent entity is its equity nexus with another identified fraudster, `s2'. When considering node features, the three most significant attributes include `Behavioral Score,' `Accumulated Released Amount', and `Overdue principal balance'. According to the expert analysis from ICBC, a behavioral score of 65 is considered relatively low, indicating a high risk. Furthermore, examining the other two features reveals that `t1' has already received a substantial accumulated released amount. However, the overdue balance generated constitutes a significant proportion compared to the accumulated released amount. This combination of factors further substantiates the classification of `t1' as fraudulent. In the second example, the classification of `t2' as a fraudulent entity is primarily predicated on the `Guarantor' edge from `S2', which is already identified as a high-risk fraudster. Regarding the node features, `t2' has provided a mortgage as the method of credit guarantee associated with a relatively extended contract duration. However, it is noteworthy that `t2' also possesses a subprime loan. The combination of these factors suggests that `t2' may lack the financial capacity to repay the debt. This amalgamation of circumstances further substantiates the classification of `t2' as a fraudulent entity. These two examples demonstrate how the explanations provide insights into the reasons behind the fraudster's predictions, aiding in understanding and interpreting the model's decisions.

\subsection{Model Analysis}
\paragraph{Different GNN encoders.} While it is possible to use alternative GNNs like GCN as feature encoders, ignoring the heterogeneity may result in performance degradation. In general, the data in ICBC contains heterogeneous types of information related to various nodes and relations thus we opt to construct our model using a customized heterogeneous graph transformer , which has demonstrated its superiority in our baseline XFraud~\cite{rao2020xfraud}. To provide a thorough analysis of model design, we also conduct experiment to evaluate different kinds of GNN based encoders. As shown in Table~\ref{tab:encoder}, heterogenous GNNs (i.e. HAN~\cite{wang2019heterogeneous} and HGT~\cite{hu2020heterogeneous}) significantly outperform homogenous GNNs including GCN and GAT. HGT has achieved a higher performances as well as less time consumption than HAN.

\begin{table}[t]
    \caption{Analysis with different GNN encoders.}
    \label{tab:encoder}
    \renewcommand\arraystretch{1}
    \centering
    \small
    \begin{tabular}{l|ccc}
    \toprule
    \textbf{Dataset} & \multicolumn{3}{c}{ICBC}\\
    \midrule
    Metrics & AUC & Recall & Epoch Time\\
    \midrule
    GCN based & 63.41{\tiny$\pm$3.17} & 58.76{\tiny$\pm$2.62} & 0.11s\\
    GAT based & 94.23{\tiny$\pm$2.31} & 93.15{\tiny$\pm$1.64} & 0.17s\\
    HAN based & 98.41{\tiny$\pm$1.55} & 97.26{\tiny$\pm$1.72} & 0.23s\\
    \midrule
    HGT based(SEFraud) & \textbf{99.69{\tiny$\pm$1.27}} & \textbf{99.38{\tiny$\pm$2.31}} & 0.19s\\
    \bottomrule
    \end{tabular}
 
\end{table}

\paragraph{Hyper-parameters analysis.} We also conducted a hyper-parameter analysis on the ICBC dataset.  One of the key hyperparameters is $\lambda$ , which balances the classification loss and contrastive tripplet loss. As shown in the Tabel~\ref{tab:hyperparameters} , the performance initially improves and then gradually declines when $\lambda$ ranges from 0 to 0.5. This behavior can be attributed to the increasing weight of the triplet loss, which enhances instance discrimination but may weaken classification performance beyond a certain threshold. A similar trend is observed for another hyperparamete $\alpha$, which acts as a margin in the triplet loss. A larger margin encourages the model to distinguish important edges and node features more confidently. However, excessively large margins can make it challenging to reduce the triplet loss to zero, potentially harming the overall model performance.
\begin{table}[t]
    \caption{Hyper-parameters analysis on ICBC dataset.}
    \label{tab:hyperparameters}
    \renewcommand\arraystretch{1}
    \centering
    \small
    \begin{tabular}{l|cccccc}
    \toprule
    \textbf{Different $\lambda$} & 0 & 0.1 & 0.2 & 0.3 & 0.4 & 0.5 \\
    \midrule
    AUC    & 96.68 & 99.10 & 99.37 & \textbf{99.69} & 98.43 & 98.18 \\ 
    Recall & 96.37 & 98.79 & 99.05 & \textbf{99.38} & 96.98 & 96.38 \\
    \midrule
    \textbf{Different $\alpha$} & 0 & 0.1 & 0.2 & 0.3 & 0.4 & 0.5\\
    \midrule
    AUC    & 99.09 & \textbf{99.69} & 99.68 & 99.39 & 98.78 & 98.76\\ 
    Recall & 98.19 & \textbf{99.38} & 98.79 & 98.19 & 97.59 & 97.59 \\
    \bottomrule
    \end{tabular}
\end{table}

\subsection{Time Comparison}

Efficiency plays a critical role in industry applications, especially for financial systems requiring milliseconds response speed. For each instance, GNNExplainer needs to re-train the explanation network before generating a new explanation and thus is criticized for its low efficiency. The explanation network in PGExplainer is shared across the population of instances and can be utilized to explain new instances in the inductive setting. However, it also needs to calculate the weights of the subgraph in each explaining process. In contrast, explanation weights in SE-Mask are trained with the backbone GNN model and can be directly extracted for explaining, which only takes milliseconds. As shown in Tabel~\ref{tab:efficiency}, we compared the inference time of generating a new explanation instance with the three kinds of methods. SE-Mask achieves over 1000x speed-up on explain Tree-Cycles and Tree-Grids datasets, having a significant advantage on time efficiency. In an industry financial fraud detection system, GNNExplainer requires approximately 7 seconds to generate a reasonable explanation for a flagged fraudster. In contrast, our proposed SE-Mask accomplishes the same task in just 0.4 milliseconds.

\begin{table}[tbp]
    \caption{Inference Time (ms).}
    \label{tab:efficiency}
    \renewcommand\arraystretch{1}
    \centering
    \small
    
    \scalebox{1.0}{
    \begin{tabular}{l|c|c|c|c}
    \toprule
    \textbf{} & {BA-2motifs} & {BA-Shapes} & {Tree-Cycles} & {Tree-Grids}\\
    \midrule
    GNNExplainer & 934.72ms  & 650.60ms & 690.13ms  & 713.40ms \\
    \textit{\#speed-up} & 1x &1x & 1x & 1x\\
    \midrule
    PGExplainer & 80.13ms  & 10.92ms & 6.36ms & 6.72ms \\
    \textit{\#speed-up} & 12x & 59x & 108x & 106x  \\
    \midrule
    SE-Mask & 10.55ms & 0.96ms & 0.54ms & 0.60ms \\
     \textit{\#speed-up} & 86x & 678x & 1278x & 1189x \\
    \bottomrule
    \end{tabular}
    }
\end{table}

\subsection{Deployment and application}
 
In the study's second phase, the proposed SEFraud is further deployed in ICBC's production environment with the Mindspore service~\cite{huawei2022huawei}. The model's performance is measured based on ICBC’s real data, which is divided into four parts corresponding to four months. Each month contains approximately 86,000 nodes and 18,000 edges. The model is trained using one month’s data and tested its predictive performances and explanation effects using the following month’s data.
The feedback received from ICBC is highly positive and inspiring. During the real business practice, The average prediction AUC of SEFraud is 97\%, indicating a high level of accuracy, with a recall rate of 0.98. Additionally, the inference time taken for predicting and providing explanations for a single node averaged just at 0.4ms. Moreover, the generated explanations for the predicted fraudsters are well-received by the experts from ICBC, demonstrating the effectiveness of our approach.

\section{Conclusion}
Current practical implementations of graph fraud detection application are limited for scenarios that require both high confidence (i.e., explainable detection) and efficiency. In this paper, we introduce the SEFraud, a highly efficient self-explainable fraud detection framework. The interpretive feature mask and edge mask are integrated into customized heterogeneous graph transformer networks to enhance the detection model's prediction ability and provide explanations. And a specific contrastive triplet loss is further designed to augment the mask learning. Experimental results demonstrate the advantages of SEFraud in various fraud detection tasks and it is able to provide reasonable explanations within milliseconds. The deployment and verification of SEFraud in ICBC, one of the largest banks in China, further verify its efficiency and applicability in real industry environment.

\section{Acknowledgment}
This work was supported by National Key R\&D program of China under Grant No.2023YFC3807603, No.2021ZD0110400. We also appreciate the help of cooperation partners in this project, and all the thoughtful and insightful suggestions from reviewers. 

\bibliographystyle{ACM-Reference-Format}
\balance
\bibliography{paper}


\begin{thebibliography}{46}


\ifx \showCODEN    \undefined \def \showCODEN     #1{\unskip}     \fi
\ifx \showDOI      \undefined \def \showDOI       #1{#1}\fi
\ifx \showISBNx    \undefined \def \showISBNx     #1{\unskip}     \fi
\ifx \showISBNxiii \undefined \def \showISBNxiii  #1{\unskip}     \fi
\ifx \showISSN     \undefined \def \showISSN      #1{\unskip}     \fi
\ifx \showLCCN     \undefined \def \showLCCN      #1{\unskip}     \fi
\ifx \shownote     \undefined \def \shownote      #1{#1}          \fi
\ifx \showarticletitle \undefined \def \showarticletitle #1{#1}   \fi
\ifx \showURL      \undefined \def \showURL       {\relax}        \fi
\providecommand\bibfield[2]{#2}
\providecommand\bibinfo[2]{#2}
\providecommand\natexlab[1]{#1}
\providecommand\showeprint[2][]{arXiv:#2}

\bibitem[Branting et~al\mbox{.}(2016)]%
        {branting2016graph}
\bibfield{author}{\bibinfo{person}{L~Karl Branting}, \bibinfo{person}{Flo Reeder}, \bibinfo{person}{Jeffrey Gold}, {and} \bibinfo{person}{Timothy Champney}.} \bibinfo{year}{2016}\natexlab{}.
\newblock \showarticletitle{Graph analytics for healthcare fraud risk estimation}. In \bibinfo{booktitle}{\emph{2016 IEEE/ACM International Conference on Advances in Social Networks Analysis and Mining (ASONAM)}}. IEEE, \bibinfo{pages}{845--851}.
\newblock


\bibitem[Breuer et~al\mbox{.}(2020)]%
        {breuer2020friend}
\bibfield{author}{\bibinfo{person}{Adam Breuer}, \bibinfo{person}{Roee Eilat}, {and} \bibinfo{person}{Udi Weinsberg}.} \bibinfo{year}{2020}\natexlab{}.
\newblock \showarticletitle{Friend or faux: Graph-based early detection of fake accounts on social networks}. In \bibinfo{booktitle}{\emph{Proceedings of The Web Conference 2020}}. \bibinfo{pages}{1287--1297}.
\newblock


\bibitem[Deng et~al\mbox{.}(2022)]%
        {deng2022markov}
\bibfield{author}{\bibinfo{person}{Leyan Deng}, \bibinfo{person}{Chenwang Wu}, \bibinfo{person}{Defu Lian}, \bibinfo{person}{Yongji Wu}, {and} \bibinfo{person}{Enhong Chen}.} \bibinfo{year}{2022}\natexlab{}.
\newblock \showarticletitle{Markov-driven graph convolutional networksfor social spammer detection}.
\newblock \bibinfo{journal}{\emph{IEEE Transactions on Knowledge and Data Engineering}} (\bibinfo{year}{2022}).
\newblock


\bibitem[Dou et~al\mbox{.}(2020)]%
        {dou2020enhancing}
\bibfield{author}{\bibinfo{person}{Yingtong Dou}, \bibinfo{person}{Zhiwei Liu}, \bibinfo{person}{Li Sun}, \bibinfo{person}{Yutong Deng}, \bibinfo{person}{Hao Peng}, {and} \bibinfo{person}{Philip~S Yu}.} \bibinfo{year}{2020}\natexlab{}.
\newblock \showarticletitle{Enhancing graph neural network-based fraud detectors against camouflaged fraudsters}. In \bibinfo{booktitle}{\emph{Proceedings of the 29th ACM international conference on information \& knowledge management}}. \bibinfo{pages}{315--324}.
\newblock


\bibitem[Fey and Lenssen(2019)]%
        {Fey/Lenssen/2019pyg}
\bibfield{author}{\bibinfo{person}{Matthias Fey} {and} \bibinfo{person}{Jan~E. Lenssen}.} \bibinfo{year}{2019}\natexlab{}.
\newblock \showarticletitle{Fast Graph Representation Learning with {PyTorch Geometric}}. In \bibinfo{booktitle}{\emph{ICLR Workshop on Representation Learning on Graphs and Manifolds}}.
\newblock


\bibitem[Fu et~al\mbox{.}(2020)]%
        {fu2020magnn}
\bibfield{author}{\bibinfo{person}{Xinyu Fu}, \bibinfo{person}{Jiani Zhang}, \bibinfo{person}{Ziqiao Meng}, {and} \bibinfo{person}{Irwin King}.} \bibinfo{year}{2020}\natexlab{}.
\newblock \showarticletitle{Magnn: Metapath aggregated graph neural network for heterogeneous graph embedding}. In \bibinfo{booktitle}{\emph{Proceedings of The Web Conference 2020}}. \bibinfo{pages}{2331--2341}.
\newblock


\bibitem[Gilmer et~al\mbox{.}(2017)]%
        {gilmer2017neural}
\bibfield{author}{\bibinfo{person}{Justin Gilmer}, \bibinfo{person}{Samuel~S Schoenholz}, \bibinfo{person}{Patrick~F Riley}, \bibinfo{person}{Oriol Vinyals}, {and} \bibinfo{person}{George~E Dahl}.} \bibinfo{year}{2017}\natexlab{}.
\newblock \showarticletitle{Neural message passing for quantum chemistry}. In \bibinfo{booktitle}{\emph{International conference on machine learning}}. PMLR, \bibinfo{pages}{1263--1272}.
\newblock


\bibitem[Hamilton et~al\mbox{.}(2017)]%
        {hamilton2017inductive}
\bibfield{author}{\bibinfo{person}{William~L Hamilton}, \bibinfo{person}{Rex Ying}, {and} \bibinfo{person}{Jure Leskovec}.} \bibinfo{year}{2017}\natexlab{}.
\newblock \showarticletitle{Inductive representation learning on large graphs}. In \bibinfo{booktitle}{\emph{Proceedings of the 31st International Conference on Neural Information Processing Systems}}. \bibinfo{pages}{1025--1035}.
\newblock


\bibitem[Hilal et~al\mbox{.}(2022)]%
        {hilal2022financial}
\bibfield{author}{\bibinfo{person}{Waleed Hilal}, \bibinfo{person}{S~Andrew Gadsden}, {and} \bibinfo{person}{John Yawney}.} \bibinfo{year}{2022}\natexlab{}.
\newblock \showarticletitle{Financial fraud: a review of anomaly detection techniques and recent advances}.
\newblock \bibinfo{journal}{\emph{Expert systems With applications}}  \bibinfo{volume}{193} (\bibinfo{year}{2022}), \bibinfo{pages}{116429}.
\newblock


\bibitem[Hochreiter and Schmidhuber(1997)]%
        {hochreiter1997long}
\bibfield{author}{\bibinfo{person}{Sepp Hochreiter} {and} \bibinfo{person}{J{\"u}rgen Schmidhuber}.} \bibinfo{year}{1997}\natexlab{}.
\newblock \showarticletitle{Long short-term memory}.
\newblock \bibinfo{journal}{\emph{Neural computation}} \bibinfo{volume}{9}, \bibinfo{number}{8} (\bibinfo{year}{1997}), \bibinfo{pages}{1735--1780}.
\newblock


\bibitem[Hu et~al\mbox{.}(2020)]%
        {hu2020heterogeneous}
\bibfield{author}{\bibinfo{person}{Ziniu Hu}, \bibinfo{person}{Yuxiao Dong}, \bibinfo{person}{Kuansan Wang}, {and} \bibinfo{person}{Yizhou Sun}.} \bibinfo{year}{2020}\natexlab{}.
\newblock \showarticletitle{Heterogeneous graph transformer}. In \bibinfo{booktitle}{\emph{Proceedings of the web conference 2020}}. \bibinfo{pages}{2704--2710}.
\newblock


\bibitem[Huawei Technologies~Co.(2022)]%
        {huawei2022huawei}
\bibfield{author}{\bibinfo{person}{Ltd. Huawei Technologies~Co.}} \bibinfo{year}{2022}\natexlab{}.
\newblock \showarticletitle{Huawei MindSpore AI Development Framework}.
\newblock In \bibinfo{booktitle}{\emph{Artificial Intelligence Technology}}. \bibinfo{publisher}{Springer}, \bibinfo{pages}{137--162}.
\newblock


\bibitem[Kingma and Ba(2014)]%
        {kingma2014adam}
\bibfield{author}{\bibinfo{person}{Diederik~P Kingma} {and} \bibinfo{person}{Jimmy Ba}.} \bibinfo{year}{2014}\natexlab{}.
\newblock \showarticletitle{Adam: A method for stochastic optimization}.
\newblock \bibinfo{journal}{\emph{arXiv preprint arXiv:1412.6980}} (\bibinfo{year}{2014}).
\newblock


\bibitem[Kipf and Welling(2017)]%
        {kipf2016semi}
\bibfield{author}{\bibinfo{person}{Thomas~N Kipf} {and} \bibinfo{person}{Max Welling}.} \bibinfo{year}{2017}\natexlab{}.
\newblock \showarticletitle{Semi-supervised classification with graph convolutional networks}.
\newblock \bibinfo{journal}{\emph{International Conference on Learning Representations (ICLR)}} (\bibinfo{year}{2017}).
\newblock


\bibitem[Li et~al\mbox{.}(2020)]%
        {li2020flowscope}
\bibfield{author}{\bibinfo{person}{Xiangfeng Li}, \bibinfo{person}{Shenghua Liu}, \bibinfo{person}{Zifeng Li}, \bibinfo{person}{Xiaotian Han}, \bibinfo{person}{Chuan Shi}, \bibinfo{person}{Bryan Hooi}, \bibinfo{person}{He Huang}, {and} \bibinfo{person}{Xueqi Cheng}.} \bibinfo{year}{2020}\natexlab{}.
\newblock \showarticletitle{Flowscope: Spotting money laundering based on graphs}. In \bibinfo{booktitle}{\emph{Proceedings of the AAAI conference on artificial intelligence}}, Vol.~\bibinfo{volume}{34}. \bibinfo{pages}{4731--4738}.
\newblock


\bibitem[Liu et~al\mbox{.}(2023)]%
        {liu2023mata}
\bibfield{author}{\bibinfo{person}{Junfeng Liu}, \bibinfo{person}{Min Zhou}, \bibinfo{person}{Shuai Ma}, {and} \bibinfo{person}{Lujia Pan}.} \bibinfo{year}{2023}\natexlab{}.
\newblock \showarticletitle{MATA*: Combining Learnable Node Matching with A* Algorithm for Approximate Graph Edit Distance Computation}. In \bibinfo{booktitle}{\emph{Proceedings of the 32nd ACM International Conference on Information and Knowledge Management}}. \bibinfo{pages}{1503--1512}.
\newblock


\bibitem[Liu et~al\mbox{.}(2021)]%
        {liu2021pick}
\bibfield{author}{\bibinfo{person}{Yang Liu}, \bibinfo{person}{Xiang Ao}, \bibinfo{person}{Zidi Qin}, \bibinfo{person}{Jianfeng Chi}, \bibinfo{person}{Jinghua Feng}, \bibinfo{person}{Hao Yang}, {and} \bibinfo{person}{Qing He}.} \bibinfo{year}{2021}\natexlab{}.
\newblock \showarticletitle{Pick and choose: a GNN-based imbalanced learning approach for fraud detection}. In \bibinfo{booktitle}{\emph{Proceedings of the web conference 2021}}. \bibinfo{pages}{3168--3177}.
\newblock


\bibitem[Liu et~al\mbox{.}(2019)]%
        {liu2019geniepath}
\bibfield{author}{\bibinfo{person}{Ziqi Liu}, \bibinfo{person}{Chaochao Chen}, \bibinfo{person}{Longfei Li}, \bibinfo{person}{Jun Zhou}, \bibinfo{person}{Xiaolong Li}, \bibinfo{person}{Le Song}, {and} \bibinfo{person}{Yuan Qi}.} \bibinfo{year}{2019}\natexlab{}.
\newblock \showarticletitle{Geniepath: Graph neural networks with adaptive receptive paths}. In \bibinfo{booktitle}{\emph{Proceedings of the AAAI Conference on Artificial Intelligence}}, Vol.~\bibinfo{volume}{33}. \bibinfo{pages}{4424--4431}.
\newblock


\bibitem[Liu et~al\mbox{.}(2018)]%
        {liu2018heterogeneous}
\bibfield{author}{\bibinfo{person}{Ziqi Liu}, \bibinfo{person}{Chaochao Chen}, \bibinfo{person}{Xinxing Yang}, \bibinfo{person}{Jun Zhou}, \bibinfo{person}{Xiaolong Li}, {and} \bibinfo{person}{Le Song}.} \bibinfo{year}{2018}\natexlab{}.
\newblock \showarticletitle{Heterogeneous graph neural networks for malicious account detection}. In \bibinfo{booktitle}{\emph{Proceedings of the 27th ACM international conference on information and knowledge management}}. \bibinfo{pages}{2077--2085}.
\newblock


\bibitem[Liu et~al\mbox{.}(2020)]%
        {liu2020alleviating}
\bibfield{author}{\bibinfo{person}{Zhiwei Liu}, \bibinfo{person}{Yingtong Dou}, \bibinfo{person}{Philip~S Yu}, \bibinfo{person}{Yutong Deng}, {and} \bibinfo{person}{Hao Peng}.} \bibinfo{year}{2020}\natexlab{}.
\newblock \showarticletitle{Alleviating the inconsistency problem of applying graph neural network to fraud detection}. In \bibinfo{booktitle}{\emph{Proceedings of the 43rd international ACM SIGIR conference on research and development in information retrieval}}. \bibinfo{pages}{1569--1572}.
\newblock


\bibitem[Luo et~al\mbox{.}(2020)]%
        {luo2020parameterized}
\bibfield{author}{\bibinfo{person}{Dongsheng Luo}, \bibinfo{person}{Wei Cheng}, \bibinfo{person}{Dongkuan Xu}, \bibinfo{person}{Wenchao Yu}, \bibinfo{person}{Bo Zong}, \bibinfo{person}{Haifeng Chen}, {and} \bibinfo{person}{Xiang Zhang}.} \bibinfo{year}{2020}\natexlab{}.
\newblock \showarticletitle{Parameterized explainer for graph neural network}.
\newblock \bibinfo{journal}{\emph{Advances in neural information processing systems}}  \bibinfo{volume}{33} (\bibinfo{year}{2020}), \bibinfo{pages}{19620--19631}.
\newblock


\bibitem[Ma et~al\mbox{.}(2018)]%
        {ma2018graphrad}
\bibfield{author}{\bibinfo{person}{Jun Ma}, \bibinfo{person}{Danqing Zhang}, \bibinfo{person}{Yun Wang}, \bibinfo{person}{Yan Zhang}, {and} \bibinfo{person}{Alexey Pozdnoukhov}.} \bibinfo{year}{2018}\natexlab{}.
\newblock \showarticletitle{GraphRAD: a graph-based risky account detection system}. In \bibinfo{booktitle}{\emph{Proceedings of ACM SIGKDD conference, London, UK}}, Vol.~\bibinfo{volume}{9}.
\newblock


\bibitem[Ma et~al\mbox{.}(2021)]%
        {9565320}
\bibfield{author}{\bibinfo{person}{Xiaoxiao Ma}, \bibinfo{person}{Jia Wu}, \bibinfo{person}{Shan Xue}, \bibinfo{person}{Jian Yang}, \bibinfo{person}{Chuan Zhou}, \bibinfo{person}{Quan~Z. Sheng}, \bibinfo{person}{Hui Xiong}, {and} \bibinfo{person}{Leman Akoglu}.} \bibinfo{year}{2021}\natexlab{}.
\newblock \showarticletitle{A Comprehensive Survey on Graph Anomaly Detection with Deep Learning}.
\newblock \bibinfo{journal}{\emph{IEEE Transactions on Knowledge and Data Engineering}} (\bibinfo{year}{2021}), \bibinfo{pages}{1--1}.
\newblock
\urldef\tempurl%
\url{https://doi.org/10.1109/TKDE.2021.3118815}
\showDOI{\tempurl}


\bibitem[Pourhabibi et~al\mbox{.}(2020)]%
        {pourhabibi2020fraud}
\bibfield{author}{\bibinfo{person}{Tahereh Pourhabibi}, \bibinfo{person}{Kok-Leong Ong}, \bibinfo{person}{Booi~H Kam}, {and} \bibinfo{person}{Yee~Ling Boo}.} \bibinfo{year}{2020}\natexlab{}.
\newblock \showarticletitle{Fraud detection: A systematic literature review of graph-based anomaly detection approaches}.
\newblock \bibinfo{journal}{\emph{Decision Support Systems}}  \bibinfo{volume}{133} (\bibinfo{year}{2020}), \bibinfo{pages}{113303}.
\newblock


\bibitem[Rao et~al\mbox{.}(2020)]%
        {rao2020xfraud}
\bibfield{author}{\bibinfo{person}{Susie~Xi Rao}, \bibinfo{person}{Shuai Zhang}, \bibinfo{person}{Zhichao Han}, \bibinfo{person}{Zitao Zhang}, \bibinfo{person}{Wei Min}, \bibinfo{person}{Zhiyao Chen}, \bibinfo{person}{Yinan Shan}, \bibinfo{person}{Yang Zhao}, {and} \bibinfo{person}{Ce Zhang}.} \bibinfo{year}{2020}\natexlab{}.
\newblock \showarticletitle{xFraud: explainable fraud transaction detection}.
\newblock \bibinfo{journal}{\emph{arXiv preprint arXiv:2011.12193}} (\bibinfo{year}{2020}).
\newblock


\bibitem[Rayana and Akoglu(2015)]%
        {rayana2015collective}
\bibfield{author}{\bibinfo{person}{Shebuti Rayana} {and} \bibinfo{person}{Leman Akoglu}.} \bibinfo{year}{2015}\natexlab{}.
\newblock \showarticletitle{Collective opinion spam detection: Bridging review networks and metadata}. In \bibinfo{booktitle}{\emph{Proceedings of the 21th acm sigkdd international conference on knowledge discovery and data mining}}. \bibinfo{pages}{985--994}.
\newblock


\bibitem[Schlichtkrull et~al\mbox{.}(2018)]%
        {schlichtkrull2018modeling}
\bibfield{author}{\bibinfo{person}{Michael Schlichtkrull}, \bibinfo{person}{Thomas~N Kipf}, \bibinfo{person}{Peter Bloem}, \bibinfo{person}{Rianne Van Den~Berg}, \bibinfo{person}{Ivan Titov}, {and} \bibinfo{person}{Max Welling}.} \bibinfo{year}{2018}\natexlab{}.
\newblock \showarticletitle{Modeling relational data with graph convolutional networks}. In \bibinfo{booktitle}{\emph{The Semantic Web: 15th International Conference, ESWC 2018, Heraklion, Crete, Greece, June 3--7, 2018, Proceedings 15}}. Springer, \bibinfo{pages}{593--607}.
\newblock


\bibitem[Song et~al\mbox{.}(2023)]%
        {song2023mm}
\bibfield{author}{\bibinfo{person}{Xuemeng Song}, \bibinfo{person}{Chun Wang}, \bibinfo{person}{Changchang Sun}, \bibinfo{person}{Shanshan Feng}, \bibinfo{person}{Min Zhou}, {and} \bibinfo{person}{Liqiang Nie}.} \bibinfo{year}{2023}\natexlab{}.
\newblock \showarticletitle{MM-FRec: Multi-Modal Enhanced Fashion Item Recommendation}.
\newblock \bibinfo{journal}{\emph{IEEE Transactions on Knowledge and Data Engineering}} (\bibinfo{year}{2023}).
\newblock


\bibitem[Taher(2021)]%
        {taher2021commerce}
\bibfield{author}{\bibinfo{person}{Ghada Taher}.} \bibinfo{year}{2021}\natexlab{}.
\newblock \showarticletitle{E-commerce: advantages and limitations}.
\newblock \bibinfo{journal}{\emph{International Journal of Academic Research in Accounting Finance and Management Sciences}} \bibinfo{volume}{11}, \bibinfo{number}{1} (\bibinfo{year}{2021}), \bibinfo{pages}{153--165}.
\newblock


\bibitem[Vaswani et~al\mbox{.}(2017)]%
        {vaswani2017attention}
\bibfield{author}{\bibinfo{person}{Ashish Vaswani}, \bibinfo{person}{Noam Shazeer}, \bibinfo{person}{Niki Parmar}, \bibinfo{person}{Jakob Uszkoreit}, \bibinfo{person}{Llion Jones}, \bibinfo{person}{Aidan~N Gomez}, \bibinfo{person}{{\L}ukasz Kaiser}, {and} \bibinfo{person}{Illia Polosukhin}.} \bibinfo{year}{2017}\natexlab{}.
\newblock \showarticletitle{Attention is all you need}.
\newblock \bibinfo{journal}{\emph{Advances in neural information processing systems}}  \bibinfo{volume}{30} (\bibinfo{year}{2017}).
\newblock


\bibitem[Veli{\v{c}}kovi{\'c} et~al\mbox{.}(2017)]%
        {velivckovic2017graph}
\bibfield{author}{\bibinfo{person}{Petar Veli{\v{c}}kovi{\'c}}, \bibinfo{person}{Guillem Cucurull}, \bibinfo{person}{Arantxa Casanova}, \bibinfo{person}{Adriana Romero}, \bibinfo{person}{Pietro Lio}, {and} \bibinfo{person}{Yoshua Bengio}.} \bibinfo{year}{2017}\natexlab{}.
\newblock \showarticletitle{Graph attention networks}.
\newblock \bibinfo{journal}{\emph{International Conference on Learning Representations (ICLR)}} (\bibinfo{year}{2017}).
\newblock


\bibitem[Wang et~al\mbox{.}(2019b)]%
        {wang2019semi}
\bibfield{author}{\bibinfo{person}{Daixin Wang}, \bibinfo{person}{Jianbin Lin}, \bibinfo{person}{Peng Cui}, \bibinfo{person}{Quanhui Jia}, \bibinfo{person}{Zhen Wang}, \bibinfo{person}{Yanming Fang}, \bibinfo{person}{Quan Yu}, \bibinfo{person}{Jun Zhou}, \bibinfo{person}{Shuang Yang}, {and} \bibinfo{person}{Yuan Qi}.} \bibinfo{year}{2019}\natexlab{b}.
\newblock \showarticletitle{A semi-supervised graph attentive network for financial fraud detection}. In \bibinfo{booktitle}{\emph{2019 IEEE International Conference on Data Mining (ICDM)}}. IEEE, \bibinfo{pages}{598--607}.
\newblock


\bibitem[Wang et~al\mbox{.}(2019c)]%
        {wang2019fdgars}
\bibfield{author}{\bibinfo{person}{Jianyu Wang}, \bibinfo{person}{Rui Wen}, \bibinfo{person}{Chunming Wu}, \bibinfo{person}{Yu Huang}, {and} \bibinfo{person}{Jian Xiong}.} \bibinfo{year}{2019}\natexlab{c}.
\newblock \showarticletitle{Fdgars: Fraudster detection via graph convolutional networks in online app review system}. In \bibinfo{booktitle}{\emph{Companion proceedings of the 2019 World Wide Web conference}}. \bibinfo{pages}{310--316}.
\newblock


\bibitem[Wang et~al\mbox{.}(2019a)]%
        {wang2019heterogeneous}
\bibfield{author}{\bibinfo{person}{Xiao Wang}, \bibinfo{person}{Houye Ji}, \bibinfo{person}{Chuan Shi}, \bibinfo{person}{Bai Wang}, \bibinfo{person}{Yanfang Ye}, \bibinfo{person}{Peng Cui}, {and} \bibinfo{person}{Philip~S Yu}.} \bibinfo{year}{2019}\natexlab{a}.
\newblock \showarticletitle{Heterogeneous graph attention network}. In \bibinfo{booktitle}{\emph{The world wide web conference}}. \bibinfo{pages}{2022--2032}.
\newblock


\bibitem[Weber et~al\mbox{.}(2019)]%
        {weber2019anti}
\bibfield{author}{\bibinfo{person}{Mark Weber}, \bibinfo{person}{Giacomo Domeniconi}, \bibinfo{person}{Jie Chen}, \bibinfo{person}{Daniel Karl~I Weidele}, \bibinfo{person}{Claudio Bellei}, \bibinfo{person}{Tom Robinson}, {and} \bibinfo{person}{Charles~E Leiserson}.} \bibinfo{year}{2019}\natexlab{}.
\newblock \showarticletitle{Anti-money laundering in bitcoin: Experimenting with graph convolutional networks for financial forensics}.
\newblock \bibinfo{journal}{\emph{arXiv preprint arXiv:1908.02591}} (\bibinfo{year}{2019}).
\newblock


\bibitem[Yang et~al\mbox{.}(2021)]%
        {yang2021discrete}
\bibfield{author}{\bibinfo{person}{Menglin Yang}, \bibinfo{person}{Min Zhou}, \bibinfo{person}{Marcus Kalander}, \bibinfo{person}{Zengfeng Huang}, {and} \bibinfo{person}{Irwin King}.} \bibinfo{year}{2021}\natexlab{}.
\newblock \showarticletitle{Discrete-time temporal network embedding via implicit hierarchical learning in hyperbolic space}. In \bibinfo{booktitle}{\emph{Proceedings of the 27th ACM SIGKDD Conference on Knowledge Discovery \& Data Mining}}. \bibinfo{pages}{1975--1985}.
\newblock


\bibitem[Yang et~al\mbox{.}(2022b)]%
        {yang2022hrcf}
\bibfield{author}{\bibinfo{person}{Menglin Yang}, \bibinfo{person}{Min Zhou}, \bibinfo{person}{Jiahong Liu}, \bibinfo{person}{Defu Lian}, {and} \bibinfo{person}{Irwin King}.} \bibinfo{year}{2022}\natexlab{b}.
\newblock \showarticletitle{{HRCF}: Enhancing Collaborative Filtering via Hyperbolic Geometric Regularization}. In \bibinfo{booktitle}{\emph{Proceedings of the Web Conference}}.
\newblock


\bibitem[Yang et~al\mbox{.}(2022a)]%
        {yang2022graph}
\bibfield{author}{\bibinfo{person}{Tianmeng Yang}, \bibinfo{person}{Yujing Wang}, \bibinfo{person}{Zhihan Yue}, \bibinfo{person}{Yaming Yang}, \bibinfo{person}{Yunhai Tong}, {and} \bibinfo{person}{Jing Bai}.} \bibinfo{year}{2022}\natexlab{a}.
\newblock \showarticletitle{Graph pointer neural networks}. In \bibinfo{booktitle}{\emph{Proceedings of the AAAI conference on artificial intelligence}}, Vol.~\bibinfo{volume}{36}. \bibinfo{pages}{8832--8839}.
\newblock


\bibitem[Yang et~al\mbox{.}(2023b)]%
        {yang2023mitigating}
\bibfield{author}{\bibinfo{person}{Tianmeng Yang}, \bibinfo{person}{Min Zhou}, \bibinfo{person}{Yujing Wang}, \bibinfo{person}{Zhengjie Lin}, \bibinfo{person}{Lujia Pan}, \bibinfo{person}{Bin Cui}, {and} \bibinfo{person}{Yunhai Tong}.} \bibinfo{year}{2023}\natexlab{b}.
\newblock \showarticletitle{Mitigating Semantic Confusion from Hostile Neighborhood for Graph Active Learning}.
\newblock \bibinfo{journal}{\emph{arXiv preprint arXiv:2308.08823}} (\bibinfo{year}{2023}).
\newblock


\bibitem[Yang et~al\mbox{.}(2023a)]%
        {yang2023simple}
\bibfield{author}{\bibinfo{person}{Xiaocheng Yang}, \bibinfo{person}{Mingyu Yan}, \bibinfo{person}{Shirui Pan}, \bibinfo{person}{Xiaochun Ye}, {and} \bibinfo{person}{Dongrui Fan}.} \bibinfo{year}{2023}\natexlab{a}.
\newblock \showarticletitle{Simple and efficient heterogeneous graph neural network}. In \bibinfo{booktitle}{\emph{Proceedings of the AAAI Conference on Artificial Intelligence}}, Vol.~\bibinfo{volume}{37}. \bibinfo{pages}{10816--10824}.
\newblock


\bibitem[Ying et~al\mbox{.}(2018)]%
        {ying2018graph}
\bibfield{author}{\bibinfo{person}{Rex Ying}, \bibinfo{person}{Ruining He}, \bibinfo{person}{Kaifeng Chen}, \bibinfo{person}{Pong Eksombatchai}, \bibinfo{person}{William~L Hamilton}, {and} \bibinfo{person}{Jure Leskovec}.} \bibinfo{year}{2018}\natexlab{}.
\newblock \showarticletitle{Graph convolutional neural networks for web-scale recommender systems}. In \bibinfo{booktitle}{\emph{Proceedings of the 24th ACM SIGKDD international conference on knowledge discovery \& data mining}}. \bibinfo{pages}{974--983}.
\newblock


\bibitem[Ying et~al\mbox{.}(2019)]%
        {ying2019gnnexplainer}
\bibfield{author}{\bibinfo{person}{Zhitao Ying}, \bibinfo{person}{Dylan Bourgeois}, \bibinfo{person}{Jiaxuan You}, \bibinfo{person}{Marinka Zitnik}, {and} \bibinfo{person}{Jure Leskovec}.} \bibinfo{year}{2019}\natexlab{}.
\newblock \showarticletitle{Gnnexplainer: Generating explanations for graph neural networks}.
\newblock \bibinfo{journal}{\emph{Advances in neural information processing systems}}  \bibinfo{volume}{32} (\bibinfo{year}{2019}).
\newblock


\bibitem[Yuan et~al\mbox{.}(2020)]%
        {yuan2020xgnn}
\bibfield{author}{\bibinfo{person}{Hao Yuan}, \bibinfo{person}{Jiliang Tang}, \bibinfo{person}{Xia Hu}, {and} \bibinfo{person}{Shuiwang Ji}.} \bibinfo{year}{2020}\natexlab{}.
\newblock \showarticletitle{Xgnn: Towards model-level explanations of graph neural networks}. In \bibinfo{booktitle}{\emph{Proceedings of the 26th ACM SIGKDD International Conference on Knowledge Discovery \& Data Mining}}. \bibinfo{pages}{430--438}.
\newblock


\bibitem[Zhang et~al\mbox{.}(2020)]%
        {zhang2020gcn}
\bibfield{author}{\bibinfo{person}{Shijie Zhang}, \bibinfo{person}{Hongzhi Yin}, \bibinfo{person}{Tong Chen}, \bibinfo{person}{Quoc Viet~Nguyen Hung}, \bibinfo{person}{Zi Huang}, {and} \bibinfo{person}{Lizhen Cui}.} \bibinfo{year}{2020}\natexlab{}.
\newblock \showarticletitle{Gcn-based user representation learning for unifying robust recommendation and fraudster detection}. In \bibinfo{booktitle}{\emph{Proceedings of the 43rd international ACM SIGIR conference on research and development in information retrieval}}. \bibinfo{pages}{689--698}.
\newblock


\bibitem[Zheng et~al\mbox{.}(2017)]%
        {zheng2017smoke}
\bibfield{author}{\bibinfo{person}{Haizhong Zheng}, \bibinfo{person}{Minhui Xue}, \bibinfo{person}{Hao Lu}, \bibinfo{person}{Shuang Hao}, \bibinfo{person}{Haojin Zhu}, \bibinfo{person}{Xiaohui Liang}, {and} \bibinfo{person}{Keith Ross}.} \bibinfo{year}{2017}\natexlab{}.
\newblock \showarticletitle{Smoke screener or straight shooter: Detecting elite sybil attacks in user-review social networks}.
\newblock \bibinfo{journal}{\emph{arXiv preprint arXiv:1709.06916}} (\bibinfo{year}{2017}).
\newblock


\bibitem[Zhong et~al\mbox{.}(2020)]%
        {zhong2020financial}
\bibfield{author}{\bibinfo{person}{Qiwei Zhong}, \bibinfo{person}{Yang Liu}, \bibinfo{person}{Xiang Ao}, \bibinfo{person}{Binbin Hu}, \bibinfo{person}{Jinghua Feng}, \bibinfo{person}{Jiayu Tang}, {and} \bibinfo{person}{Qing He}.} \bibinfo{year}{2020}\natexlab{}.
\newblock \showarticletitle{Financial defaulter detection on online credit payment via multi-view attributed heterogeneous information network}. In \bibinfo{booktitle}{\emph{Proceedings of The Web Conference 2020}}. \bibinfo{pages}{785--795}.
\newblock


\end{thebibliography}

\appendix
\nobalance


\end{document}